\documentclass[11pt]{article}

\usepackage[final]{acl}

\usepackage{times}
\usepackage{latexsym}

\usepackage[T1]{fontenc}

\usepackage[utf8]{inputenc}

\usepackage{microtype}

\usepackage{inconsolata}

\usepackage{graphicx}
\usepackage[table]{xcolor}
\usepackage{pifont}
\usepackage{bbding}
\usepackage{algorithm}
\usepackage{algorithmic}
\usepackage{booktabs}
\usepackage{multirow}
\usepackage{multicol}
\usepackage{amsmath, amssymb, bm}
\usepackage{mathtools, physics}
\usepackage[most]{tcolorbox}
\usepackage{arydshln}

\title{LAG: Logic-Augmented Generation from a Cartesian Perspective}

\author{Yilin Xiao, Chuang Zhou, Yujing Zhang, Qinggang Zhang \\ {\bf Su Dong, Shengyuan Chen, Chang Yang, Xiao Huang } \\
        The Hong Kong Polytechnic University \\ Hong Kong SAR \\ \texttt{yilin.xiao@connect.polyu.hk, \{qinggang.zhang, xiao.huang\}@polyu.edu.hk}}

\begin{document}
\maketitle
\begin{abstract}
Large language models (LLMs) have demonstrated remarkable capabilities across a wide range of tasks, yet exhibit critical limitations in knowledge-intensive tasks, often generating hallucinations when faced with questions requiring specialized expertise. While retrieval-augmented generation (RAG) mitigates this by integrating external knowledge, it struggles with complex reasoning scenarios due to its reliance on direct semantic retrieval and lack of structured logical organization. Inspired by Cartesian principles from \textit{Discours de la méthode}, this paper introduces Logic-Augmented Generation (LAG), a novel paradigm that reframes knowledge augmentation through systematic question decomposition, atomic memory bank and logic-aware reasoning. Specifically, LAG first decomposes complex questions into atomic sub-questions ordered by logical dependencies. It then resolves these sequentially, using prior answers to guide context retrieval for subsequent sub-questions, ensuring stepwise grounding in the logical chain. Experiments on four benchmarks demonstrate that LAG significantly improves accuracy and reduces hallucination over existing methods. 

\end{abstract}

\section{Introduction}
Large language models (LLMs), like  Claude~\cite{anthropic2024claude}, ChatGPT~\cite{openai2023gpt4} and the Deepseek series~\cite{liu2024deepseek}, have demonstrated remarkable capabilities in many real-world tasks, such as question answering~\cite{allam2012question}, text comprehension~\cite{wright2017systematic} and content generation~\cite{kumar2024large}. Despite the success, these models are often criticized for their tendency to produce hallucinations, generating incorrect statements on tasks beyond their perception~\cite{ji2023towards,zhang2024knowgpt}. Recently, retrieval-augmented generation (RAG)~\cite{gao2023retrieval,lewis2020retrieval} has emerged as a promising solution to alleviate such hallucinations.
By dynamically leveraging external knowledge from textual corpora,
RAG enables LLMs to generate more accurate and reliable responses without costly retraining~\cite{lewis2020retrieval,devalal2018lora}.

\begin{figure}[t]
    \centering
    \includegraphics[width=1\linewidth]{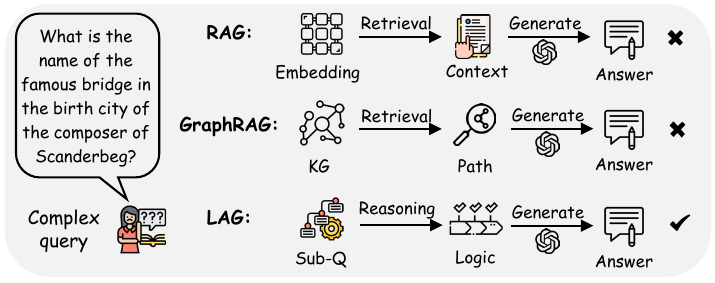}
    \caption{Comparison of three RAG paradigms. LAG offers a superior balance of efficiency and accuracy, providing a more lightweight solution than GraphRAG while outperforming it and traditional RAG in accuracy.}
    \label{fig:example}
\end{figure}

RAG systems typically operate through three key stages: knowledge preprocessing, retrieval, and integration. First, external textual corpora are segmented into manageable chunks and converted into vector representations for efficient indexing. When a query is received, the system then retrieves relevant text segments using semantic similarity matching~\cite{sawarkar2024blended} or keyword-based search~\cite{purwar2023keyword}. Finally, during integration, the retrieved information is combined with the original query to produce knowledge-enhanced responses. Recent advances in RAG technology have evolved beyond basic text retrieval toward more sophisticated approaches. These include graph-based systems~\cite{zhang2025survey,peng2024graph,procko2024graph} that model conceptual relationships using graph structures, hierarchical methods~\cite{chen2024multi,li2024structrag} preserving document organization through multi-level retrieval, re-ranking implementations~\cite{glass2022re2g,xu2023recomp} utilizing preliminary retrieval followed by refined scoring, Self-RAG architectures~\cite{asai2023self} capable of on-demand retrieval and self-reflection, and adaptive frameworks~\cite{tang2024mba,raptor} that dynamically adjust retrieval strategies based on query complexity. These strategies significantly enhance naive RAG systems through improved retrieval accuracy. 

However, despite the potential of this retrieval-centric architecture, existing RAG systems exhibit three critical limitations when handling questions of high complexity. \ding{182} Direct retrieval using semantic or keyword matching often fails to capture the underlying logical structure of complex questions, leading to irrelevant or fragmented context. For instance, retrieving with the question shown in Figure~\ref{fig:example} returns only information about \texttt{Scanderbeg}, which is insufficient to arrive at the correct answer. \ding{183} When relevant knowledge is retrieved, RAG lacks mechanisms to organize information according to inherent logical dependencies, limiting coherent reasoning in practical scenarios. Revisiting the question in Figure~\ref{fig:example}, even when relevant context is retrieved, the LLM still often struggle because it fails to capture the logical dependencies inherent (\texttt{Scanderbeg$\to$composer$\to$birth city$\to$famous bridge}) in the question. \ding{184} Although several methods~\cite{li2025cot,ircot} use the Chain-of-Thought~\cite{cot} to assist in retrieval or reasoning, the overall process remains uncontrolled. These methods mainly rely on the semantic capabilities of LLMs, often resulting in unstable reasoning chains, where initial errors can be irreversibly propagated. These gaps reveal a fundamental misalignment with human cognitive processes, where problem-solving involves systematic decomposition and controllable reasoning rather than brute-force retrieval.

To bridge this gap, we introduce Logic-Augmented Generation (LAG), a novel paradigm inspired by Cartesian principles outlined in \textit{Discours de la méthode}. LAG introduces a new reasoning-first pipeline that integrates systematic decomposition, atomic memory bank and controllable reasoning into retrieval-augmented generation. Instead of immediately invoking a retriever, LAG begins by carefully analyzing the question and breaking it down into a set of atomic sub-questions that follow a logical dependency structure. To enhance efficiency, LAG utilizes an atomic memory bank, which stores and retrieves cached, high-confidence solutions to recurrent atomic sub-questions. The system then answers these sub-questions step by step, first checking the memory bank for verified solutions to avoid redundancy and potential hallucinations. When a novel sub-question is encountered, it is resolved from first principles.
This process starts with the most basic, independent sub-questions. As each sub-question is resolved, its answer becomes part of the context used to guide the retrieval and resolution of the next, more dependent sub-question. The final answer is synthesized only after all necessary sub-questions have been addressed. If an inconsistency arises during reasoning, the logical terminator triggers the activation of the alternative solution. Our main contribution is listed as follows:


\begin{itemize}
    \item We identify key limitations of RAG, and propose LAG, a reasoning-first framework that integrates systematic decomposition, atomic memory bank and logical reasoning.

    \item LAG decomposes questions into logically-dependent sub-questions and resolves them sequentially, first retrieving verified solutions from the memory bank, then solving novel sub-questions following the logical structure.
    
    \item To prevent error propagation, LAG incorporates a logical termination mechanism that halts inference upon encountering unreasonable situations. 
    
    \item Extensive experiments demonstrate that LAG significantly enhances reasoning robustness, reduces hallucination, and aligns LLM problem-solving with structured human cognition, offering a principled alternative to conventional RAG systems.
\end{itemize}

\begin{figure*}[t]
    \centering
    \includegraphics[width=1.0\linewidth]{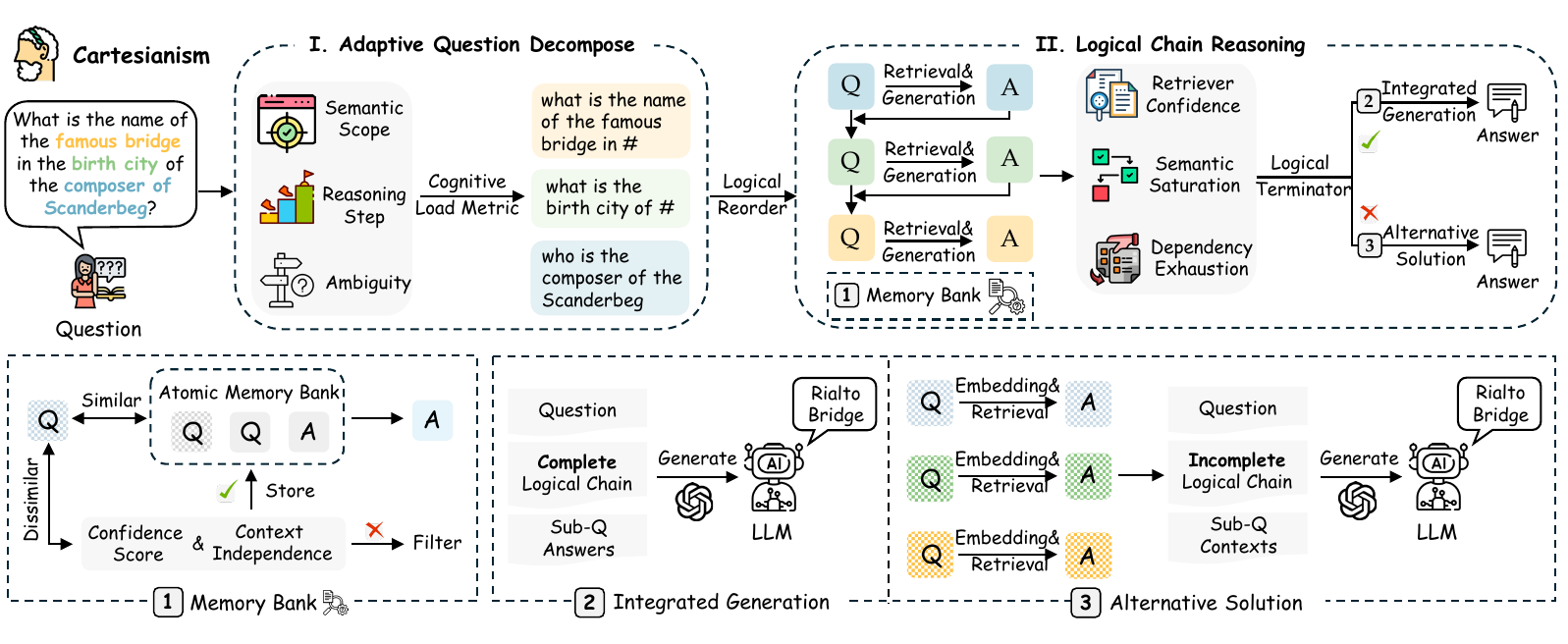}
    \caption{The framework of \textbf{LAG}. (I) Adaptive question decomposition splits complex queries into atomic sub-questions using a cognitive load. (II) Logical chain reasoning resolves sub-Q by the order of logical dependency, utilizing an atomic memory bank for recurrent knowledge. Logical terminator halts unreliable chains early. Finally, answers are synthesized via integrated generation (complete chains) or alternative solution (terminated chains).}
    \label{fig:framework}
\end{figure*}


\section{Related Work}
RAG has emerged as a critical framework for enhancing LLMs by integrating external knowledge. Early approaches such as REALM~\cite{REALM} and DPR~\cite{DPR} focus on encoding large text corpora into dense embeddings. In recent years, GraphRAG has become a new direction because it can structure fragmented knowledge. RAPTOR~\cite{raptor} and Microsoft’s GraphRAG~\cite{graphrag} both use hierarchical clustering: RAPTOR constructs recursive trees with multi-level summarization, and GraphRAG applies community detection with LLM-generated synopses, to support coarse-to-fine retrieval and high-coverage responses. DALK~\cite{dalk} and KGP~\cite{KGP} introduce dynamic KG construction and traversal agents, using LLMs to build domain-specific graphs and self-aware retrieval policies, to inject structural context while reducing noise. GFM-RAG~\cite{GFM}, G-Retriever~\cite{G-retriever}, and LightRAG~\cite{lightrag} combine graph neural encoders with specialized retrieval objectives: a query-dependent GNN trained in two stages for multi-hop generalizability, a Prize Collecting Steiner Tree formulation to reduce hallucination and improve scalability, and a dual-level graph-augmented index for efficient, incrementally updatable lookup, respectively enabling accurate, scalable reasoning over graphs. HippoRAG~\cite{hipporag}, inspired by hippocampal memory processes, leverages Personalized PageRank for single-step multi-hop retrieval, delivering state-of-the-art efficiency and performance on both path following and path finding QA tasks. HippoRAG2~\cite{hippo2} further optimizes knowledge graph refinement and deeper passage integration. More details of related work are provided in the Appendix.

\section{Preliminaries.}  
Retrieval-Augmented Generation (RAG) enhances language models by incorporating external knowledge retrieved from a large corpus. We denote the input as a natural language question $q$, which may involve latent constraints, or multi-hop reasoning. The system has access to a retrieval corpus $\mathcal{C} = \{c_1, c_2, \dots, c_N\}$, where each $c_i$ represents a passage, document chunk, or knowledge entry consisting of unstructured text. These entries may vary in granularity and source (e.g., Wikipedia, scientific papers, web documents), but are assumed to be independently indexable and retrievable. Given a query $q$, a retriever $\mathcal{R}$ returns a ranked list of relevant passages $\mathcal{R}(q) \subset \mathcal{C}$ to support downstream reasoning. Each retrieved item $c \in \mathcal{C}$ is treated as a semantically self-contained unit of information, which the system uses as external evidence during the generation or verification process.

\section{The Framework of LAG}
In \textit{Discours de la méthode}, Descartes proposed four principles for solving problems scientifically: (i) Doubt everything, avoiding precipitancy and prejudice. (ii) Divide any complex question into multiple simpler subquestions. (iii) Order sub‑questions from the simplest to the most complex and fix them step by step. (iv) Once all issues are solved, review them to ensure that nothing was omitted. Inspired by this principle, LAG introduces a novel reasoning‑first paradigm that directly aligns with it. First, to avoid precipitancy, LAG does not perform direct retrieval of the entire question. Second, the adaptive decomposition module decomposes the complex query into multiple atomic sub-questions. Third, the logical reorder module arranges these sub-questions according to their logical dependency, and the logical chain reasoning module resolves them accordingly. Finally, final answers are constructed by logically combining all sub-solutions, followed by validation against the original question to ensure complete coverage. Notably, to enhance both efficiency and robustness, LAG incorporates two key mechanisms into its reasoning pipeline: (i) a \textbf{logic-guided reasoning strategy} that leverages resolved sub-questions to guide subsequent retrieval and answering, and (ii) an \textbf{insurance mechanism} where a logical terminator is activated to initiate the alternative solution pathway if the reasoning process is deemed invalid.

\subsection{Adaptive Question Decomposition}

Our decomposition module employs cognitive load to dynamically split complex questions into verifiable atomic sub-questions. Such a mechanism decomposes complex queries through a recursive doubt-and-verify process, as exemplified by the question \textit{``What is the name of the famous bridge in the birth city of the composer of Scanderbeg?''}. While traditional retrieval systems might directly search for context related to Scanderbeg, potentially confusing \textit{``where is the bridge?''}, our method first generates verified sub-questions: [\textit{``1. who is the composer of the Scanderbeg?''}, \textit{``2. what is the birth city of \#?''}, \textit{``3. what is the name of the famous bridge in \#?''}]. The process combines cognitive load estimation with recursive refinement:

\begin{equation}\small
\text{SplitCondition}(q) = \begin{cases}
\text{True} & \text{if } \mathrm{CL}(q) > \tau(t) \\
\text{False} & \text{otherwise}
\end{cases}
\end{equation}

where the \textit{Cognitive Load} metric comprises:

{\small\[
\mathrm{CL}(q) = \underbrace{\sigma(\mathrm{Var}(\phi(q)))}_{\textsc{Semantic Scope}} + \underbrace{\sigma(\mathrm{Depth}(q))}_{\textsc{Reasoning Steps}} + \underbrace{\sigma(\mathcal{H}(q))}_{\textsc{Ambiguity}}
\]}

The Cognitive Load metric $\mathrm{CL}(q)$ integrates three complementary signals to estimate the complexity of a question. Firstly, \textbf{Semantic Scope}, is computed as the variance of the question’s embedding $\phi(q)$, capturing how broad the question's semantic coverage is. A higher value often indicates a wider topic range or more entangled concepts. The second term, \textbf{Reasoning Steps}, measures the depth of compositional reasoning required to answer $q$. LLMs estimate this by counting the number of latent inference steps involved. Lastly, \textbf{Ambiguity}, quantifies semantic uncertainty through a heuristic entropy-based function $\mathcal{H}(q)$, which reflects referential ambiguity (Details are in the appendix). $\sigma(\cdot)$ represents normalization function. Once $\mathrm{CL}(q)$ exceeds the threshold $\tau(t)$, which decays over time to encourage early resolution, our module recursively fractures $q$ into smaller sub-questions until all resulting $q_i$ satisfy $\mathrm{CL}(q_i) \leq \tau(t)$. This recursive refinement balances the need for logical soundness and factual verifiability with the goal of minimizing unnecessary conversational turns.

\subsection{Logical Chain Reasoning}

The third rule in Cartesian principles teaches us to solve problems by starting with the simplest parts and gradually working up to the more complicated ones. This mirrors how humans naturally reason: we first establish what we know with certainty, then turn to the more challenging questions. Similarly, our LAG system breaks down questions into smaller parts and solves them. Before finalizing the reasoning order, the system analyzes all decomposed questions to identify their logical relationships. This reordering rearranges that basic factual questions form the foundation, followed by analytical or comparative ones. Next, we solve questions in a logical order and use the logical information of the preceding questions to guide the retrieval of subsequent questions. At every step, three safeguards ensure reliability: 1)Is the system confident in its response? 2)Does this answer make sense with what came before? 3)Does it have enough good information? If any of these checks fail, the system knows to stop rather than guessing. In empirical evaluations, this structured, self-verifying strategy not only outputs more interpretable reasoning traces but also strengthens the justifications for its conclusions.

\subsubsection{Logic-Guided Retrieval}

After prior sub-question is answered, we update the retrieval query by incorporating both its corresponding answer and the subsequent sub-question into a single textual context. Instead of directly combining embeddings, we concatenate the prior answer $a_i$ and subsequent sub-question $q_{i+1}$ into a natural language form, e.g., ``$A_i: a_i$, $Q_{i+1}: q_{i+1}$'', and encode the resulting text to obtain a query vector for the next retrieval step. Formally, the query embedding at step $i+1$ is shown as:
\begin{equation}
\mathbf{q}^{(i+1)} = \phi(\texttt{concat}(a_i, q_{i+1})),
\end{equation}
where $\texttt{concat}(q_i, a_i)$ denotes the textual concatenation of the sub-question and prior answer, and $\phi(\cdot)$ is a shared encoder used for both questions and passages. The vector $\mathbf{q}^{(i+1)}$ is then used to query the corpus $\mathcal{C}$, retrieving a set of passages $\mathcal{R}(q^{(i+1)})$ to support resolution of the sub-question $q_{i+1}$. This context-aware retrieval process allows the system to progressively incorporate verified knowledge into subsequent steps, enabling more precise evidence collection along the logical chain.

\subsubsection{Logical Terminator}
To ensure both efficiency and robustness, we design an automatic stopping mechanism that prevents excessive or unnecessary reasoning during the logical chain reasoning. This component plays a critical role in avoiding error propagation from unanswerable sub-questions and unnecessary computation. By monitoring retrieval confidence, logical dependency states, and semantic redundancy, the system dynamically determines when to halt further reasoning, ensuring that the model focuses its efforts only when informative progress can be made.

\noindent \textbf{Retriever Confidence Drop}. We monitor the retriever's output across consecutive sub-questions. If the top-$k$ retrieved passages for a given sub-question all exhibit low semantic similarity with the query, the system interprets this as a signal of insufficient external support. Let $\text{sim}(\mathbf{q}', c_i)$ denote the cosine similarity between a sub-question $q'$ and a retrieved passage $c_i \in \mathcal{R}(q')$. If
\begin{equation}
\frac{1}{k} \sum_{i=1}^{k} \mathbb{I}\left[\text{sim}(\mathbf{q}', c_i) < \delta\right] = 1,
\end{equation}
where $\delta$ is a pre-defined similarity threshold (0.3), the resolution process for $q'$ is terminated early, avoiding further propagation of uncertainty.

\noindent \textbf{Dependency Exhaustion}. In our logical chain reasoning framework, sub-questions are arranged based on their logical dependencies. When all prerequisite sub-questions for query $q'$ have been successfully resolved, but the query still lacks sufficient support or a valid answer, the system considers the reasoning chain to be exhausted. Formally, let $\text{Deps}(q') = \{q_1, q_2, \dots, q_m\}$ denote the set of sub-questions that $q'$ depends on. If all $q_i \in \text{Deps}(q')$ are answered and yet $q'$ cannot be resolved, we halt further reasoning:
\begin{equation}\small
\left(\bigwedge_{i=1}^{m} \text{Answered}(q_i)\right) \wedge \neg \text{Answerable}(q') \Rightarrow \text{Stop}.
\end{equation}

\noindent \textbf{Semantic Saturation}. Beyond static step limits defined in the previous section, our framework actively monitors reasoning progress and halts retrieval based on a semantic saturation criterion. This is determined by evaluating the redundancy of new information relative to the accumulated context. Let $\mathcal{C}_{\text{prev}}$ denote the set of previously retrieved passages and $c_{\text{new}}$ a candidate passage. The framework deems the information space saturated and halts retrieval when the similarity $\text{sim}(c_{\text{new}}, \mathcal{C}_{\text{prev}}) > \gamma$ ( $\gamma$ is set to be 0.9). This prevents the system from iterating further when little new information is being uncovered.

\subsubsection{Atomic Memory Bank}
To enhance reasoning consistency and efficiency, we introduce the \textit{Atomic Memory Bank}, designed to store atomic knowledge units. By caching and reusing high-confidence solutions to atomic sub-question (generated via LAG), we mitigate the redundancy and hallucinations associated with repeatedly invoking the LLM for identical sub-tasks.

Formally, the memory bank is maintained as $\mathcal{M} = \{(q'_i, a'_i, \mathbf{q'}_i)\}_{i=1}^{|\mathcal{M}|}$, comprising tuples of a sub-question $q'_i$, its answer $a'_i$, and the vector embedding $\mathbf{q'}_i$. During the inference, for a given query sub-question $q'_*$, we compute its embedding $\mathbf{q'}_*$ and retrieve the stored answer if the similarity with an existing entry exceeds a threshold $\gamma$:

\begin{equation}
\max_i \frac{\mathbf{q'}_* \cdot \mathbf{q'}_i}{\|\mathbf{q'}_*\|\|\mathbf{q'}_i\|} > \gamma,
\end{equation}

where the threshold \( \gamma \) is set to 0.9. This value coincides with the semantic saturation threshold, establishing a unified criterion for the system's critical decisions. For the memory update process, we implement a strict filtering protocol to ensure knowledge quality. When the LLM solves a new sub-question, it assigns a confidence score on a five-point scale. Only pairs achieving the highest confidence level are candidates for storage. Furthermore, we incorporate a \textit{Context Independence Validation} to detect and filter sub-questions containing anaphoric references or other context-dependent ambiguities, which could be found at Appendix. Only sub-questions that pass both the confidence and independence checks are committed to $\mathcal{M}$.

\subsection{Integrated Generation}
As shown in Figure \ref{fig:framework}, our framework synthesizes the final answer by integrating all validated sub-question responses through a composition process. The system retrieves relevant evidence for each sub-query. Building upon the established reasoning chain, we first generate a comprehensive draft answer that incorporates each verified sub-solution while maintaining logical coherence with the original query. This draft must properly address all sub-questions without contradiction while fully covering the scope of the initial problem. When inconsistencies are identified, the logical terminator will stop further reasoning and merely retain the reliable logical chain, then proceed to the alternative solution, which will feed sub-questions, reliable logical chain, and retrieved context to the LLM to generate the final response. For unresolved sub-questions, instead of relying on prior knowledge, the information retrieved will be used as the basis.

\begin{table*}[t]
        \small
	\centering
	\setlength{\tabcolsep}{1mm}
	\begin{tabular}{lcccccc} 
	\toprule
 \multirow{2}{*}{\textbf{Method}}
    &\multicolumn{2}{c}{\textbf{HotpotQA}}
	& \multicolumn{2}{c}{\textbf{2Wiki}} 
	&\multicolumn{2}{c}{\textbf{MuSiQue}}\\
	\cmidrule(lr){2-3} \cmidrule(lr){4-5} \cmidrule(lr){6-7}
	&Contain-Acc. &GPT-Acc. &Contain-Acc. &GPT-Acc. &Contain-Acc. &GPT-Acc.
        \cr \cmidrule{1-7} 
    \multicolumn{7}{c}{\textbf{\textit{Direct Zero-shot LLM Inference}}}
    \cr \cmidrule{1-7}
        Llama3 (8B)~\cite{llama} & 23.7& 20.1 & 33.8 & 15.4 & 6.4 & 6.0\\
        GPT-3.5-turbo~\cite{gpt} & 31.5 & 35.4 & 31.0 & 29.9 & 7.9 & 10.9 \\
        GPT-4o-mini~\cite{gpt} &30.4 & 34.2 &29.0 &28.6 &7.8 &10.1
        \cr \cmidrule{1-7}
        \multicolumn{7}{c}{\textbf{\textit{Vanilla Retrieval-Augmented-Generation}}}
        \cr \cmidrule{1-7}
        Retrieval (Top-3) &52.1 & 55.1 & 45.1 & 43.1 &23.4 & 27.1\\
        Retrieval (Top-5) &54.6 & 56.8 & 46.6 & 45.3 &25.6 & 29.0\\
        Retrieval (Top-10) &56.0 & 58.6 &48.7 & 45.8 & 26.7 & 31.2\\
        CoT (Top-5)~\cite{cot}&55.1 &57.1&48.7&45.9 &27.1 & 30.7\\
        IRCoT (Top-5)~\cite{ircot}&58.4&59.6&53.0 &36.8 & 22.6 & 26.1
        \cr \midrule 
        \multicolumn{7}{c}{\textbf{\textit{Novel Graph Retrieval-Augmented-Generation}}}
        \cr \cmidrule{1-7}
        KGP~\cite{KGP}& 56.2& 57.1 & 52.2& 33.9& 30.5& 27.3 \\
        G-retriever ~\cite{G-retriever} & 41.3&40.9 & 47.8& 25.7 & 14.1 & 15.6\\ 
        RAPTOR ~\cite{raptor}& 58.1 & 55.3 & 60.6&43.9 & 32.2 &29.7 \\

        LightRAG ~\cite{lightrag} &61.5 &60.5 &54.4 &38.0 &27.7 &28.3\\
        HippoRAG (single-step)~\cite{hipporag} & 55.2 & 57.9 &63.7 &57.5 &31.4 &30.1\\
        HippoRAG (multi-step)~\cite{hipporag} & 61.1&63.6&66.4&62.4&34.0 &31.8 \\
        GFM-RAG (single-step) ~\cite{GFM} &61.4 &64.8 & 66.2 & 61.1 &29.3 & 32.6\\
        GFM-RAG (multi-step)~\cite{GFM} & 63.4 & 65.5 & 69.5 & 63.2 & 31.5 & 35.5 \\

        HippoRAG 2~\cite{hippo2} & 61.2&64.3 & 62.0&58.8 &34.5 &35.6
         
         \\
         LinearRAG ~\cite{linear} &64.2 & 64.9& 69.5& 62.6&32.8&36.9
         
         \\\midrule
       \rowcolor{gray!10} \textbf{LAG(Ours)} & \textbf{68.5}&\textbf{69.7} & \textbf{71.9}&\textbf{64.0} &\textbf{42.8} &\textbf{44.3}
        
        \cr 
        \bottomrule

	\end{tabular} 
    \caption{Performance comparison among state-of-the-art baselines and LAG on three benchmark datasets in terms of both Contain-Match and GPT-Evaluation Accuracy.}
	\label{tab:main_results}
\end{table*}

\section{Experiment}
\subsection{Experimental Setup}
\paragraph{Dataset.}To evaluate the effectiveness of LAG, we conducted experiments on three standard public multi-hop QA benchmarks: HotpotQA~\cite{hotpotqa}, MuSiQue~\cite{musi}, and 2WikiMultiHopQA (2Wiki)~\cite{2wiki}.  Following the evaluation protocol of HippoRAG, we used the same corpus for retrieval and sampled 1,000 questions(same with HippoRAG) from each validation set as our test queries. This setup ensures a fair, apples‑to‑apples comparison across methods. To further assess LAG’s reasoning capabilities, we also evaluated it on the recently released GraphRAG‑Bench~\cite{mybench}, which proves that LAG not only generate correct answers but also maintains the rationality of reasoning.

\paragraph{Baseline.}We compared our approach against a diverse set of established baselines, grouped into three categories: 1) LLM-Only: Direct question answering using a large language model without any external retrieval. 2) Vanilla RAG: Integration of semantic retrieval with chain‑of‑thought prompting to guide the LLM’s generation. 3) SOTA GraphRAG Methods: Recent, high-performing GraphRAG methods, including HippoRAG 2~\cite{hippo2}, GFM‑RAG~\cite{GFM}, LinearRAG~\cite{linear}, HippoRAG~\cite{hipporag}, LightRAG~\cite{lightrag}, KGP~\cite{KGP}, G‑Retriever~\cite{G-retriever}, and RAPTOR ~\cite{raptor}. Detailed description of baselines is in the Appendix.

\paragraph{Metrics.}Exact string matching can be overly stringent for multi-hop QA, as variations in casing, grammar, tense, or paraphrasing may cause a correct response to be marked wrong. Following existing work~\cite{linear}, we adopt two complementary metrics: 1) Contain-Match Accuracy: Measures whether the predicted answer contains the gold answer as a sub-string. This metric accommodates minor surface-form differences while still enforcing semantic correctness. 2) GPT-Evaluation Accuracy: An LLM‑based evaluation in which the model receives the question, the gold answer, and the prediction, then judges whether the prediction is semantically equivalent to the gold answer. These metrics together provide a balanced assessment of both surface‑level fidelity and deeper semantic correctness. For challenging reasoning task, we follow metric setting of the benchmark~\cite{mybench}.

\paragraph{Implementation Details.}Both our proposed method and all baselines utilize GPT-4o-mini as the default LLM. All experiments were executed on the RTX 4090 D. For all the methods, we use all-MiniLM-L6-v2 as the embedding model. For top-k parameters across methods, we set k = 5. In the Appendix, we provide additional experimental results regarding efficiency, hyperparameters, LLM backbones and comprehensiveness.

\begin{table}[h]
\small
\centering
\resizebox{\columnwidth}{!}{%
\begin{tabular}{lcc}
\toprule
\multirow{2}{*}{\textbf{Method}} & \multicolumn{2}{c}{\textbf{GraphRAG-Bench}} \\ \cmidrule{2-3}
                        & R Score     & AR Score     \\ \midrule
KGP~\cite{KGP}                     & 58.7                & 42.2         \\
G-retriever~\cite{G-retriever}             & 60.2                & 43.7         \\
LightRAG~\cite{lightrag}                & 60.5                & 43.8         \\
GFM-RAG~\cite{GFM}                 & 60.4                & 44.3         \\
HippoRAG~\cite{hipporag}                & 60.9                & 44.6         \\
HippoRAG 2~\cite{hippo2}              & 59.8                & 43.7         \\
RAPTOR~\cite{raptor}                  & 60.8                & 45.5         \\
LinearRAG~\cite{linear}              &  61.5    & 45.4    \\\midrule
\rowcolor{gray!10}\textbf{LAG(Ours)}            & \textbf{65.2}                & \textbf{46.4}         \cr \bottomrule
\end{tabular}}
\caption{Reasoning performance comparison among SOTA baselines and LAG on GraphRAG-Bench. R score is used to evaluate the consistency between the generated rationales and gold rationales. AR Score is an evaluation of generated answers based on the R score.} 
\label{resoning_task}
\end{table}


\begin{figure*}[t]
    \centering
    \includegraphics[width=\linewidth]{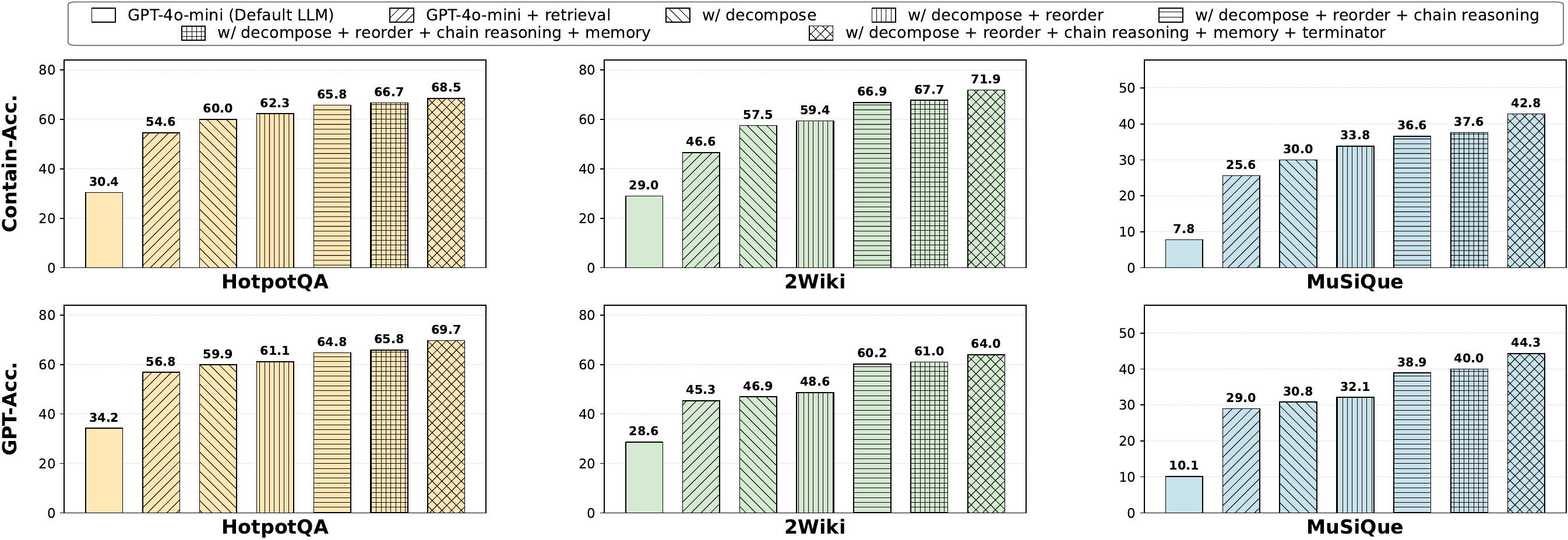}
    \caption{Ablation study of proposed LAG on three standard datasets.}
    \label{fig:ablation}
\end{figure*}

\subsection{Main Results}
As shown in Table \ref{tab:main_results}, the base LLM exhibited weak performance when directly addressing these complex questions. When simple semantic retrieval and CoT prompting were incorporated, response quality improved notably. Performance across novel RAG methods exhibits variability: KGP, RAPTOR and LightRAG demonstrated improvements in some scenarios, but they did not consistently outperform the Vanilla RAG. In contrast, HippoRAG, LinearRAG, GFM-RAG, and HippoRAG 2 consistently achieved performance gains across all datasets.

Our proposed LAG method significantly outperformed all baseline approaches on both Contain-Acc and GPT-Acc metrics. Specifically, compared to the default LLM (GPT-4o-mini), LAG achieved absolute improvements of approximately 40 points in both Contain-Acc and GPT-Acc on the HotpotQA and 2Wiki datasets, with similar gains of around 30 points observed on the MuSiQue dataset. Relative to existing RAG baselines, LAG maintained a significant advantage over most methods; even compared to the strong performers LinearRAG, HippoRAG 2 and GFM-RAG, LAG's superiority remained pronounced, particularly when handling the challenging MuSiQue dataset. Overall, these results confirm that LAG not only elevates answer accuracy across diverse domains but also ensures stable performance where other RAG approaches struggle. The marked improvements highlight LAG's superior logical capabilities in RAG.

\subsection{Challenging Reasoning Task}
Our experiments demonstrate that LAG not only achieves strong accuracy in multi-hop question answering but also excels at complex reasoning challenges. As Table ~\ref{resoning_task} shows, existing RAG methods yield reasoning scores comparable to one another, whereas LAG significantly widens this gap. This improvement arises from LAG’s explicit decomposition of a complex question into logically ordered sub-questions, followed by step‑wise solution along the resulting reasoning chain. Consequently, the rationales generated by LAG more closely align with the standard scientific explanations. Moreover, when evaluated using the AR metric, LAG again outperforms all baselines, indicating its ability to balance rigorous logical inference with accurate answer generation. Together, these results confirm that LAG substantially enhances reasoning capability over RAG systems.

\subsection{Verification of the Importance of Logic}
We hypothesize that the effectiveness of our proposed LAG derives fundamentally from the preservation of logic. To substantiate this claim, we first invoke the Cartesian principle, which establishes a theoretical foundation for the role of logic in reasoning systems. We then perform an empirical validation: in the case of a vanilla RAG, we observe that maintaining a logical order yields markedly better performance on complex tasks. To isolate the contribution of logical order within LAG, we introduce two controlled perturbations: (1) We don't concatenate the embedding of the former dependency question to the subsequent question. (2) We shuffle the order of the logical chain before retrieval and generation. Both interventions incur a statistically significant drop in performance, thereby confirming that logic is essential to RAG systems.

\begin{table}[h]
\small
\centering
\resizebox{\columnwidth}{!}{%
\begin{tabular}{lcc}
\toprule
\multirow{2}{*}{\textbf{Method}} & \multicolumn{2}{c}{\textbf{MuSiQue}} \\ \cmidrule{2-3}
                        & Contain-Acc.     & GPT-Acc.     \\ \midrule
Vanilla RAG (Random order)    &    24.4     &   27.6  \\
Vanilla RAG (Logical order)  &     27.0      & 31.8      \\ \midrule
LAG (wo/ former embedding)       &    33.5      &  36.1  \\
LAG (Random order)            &     35.1    &   36.4  \\ 
LAG (logical order)   &  \textbf{42.8}     &  \textbf{44.3}   \cr \bottomrule
\end{tabular}}
\caption{Verification of the importance of logic.} 
\label{logic Verification}
\end{table}

\subsection{Ablation Study}
To validate the effectiveness of each component in our proposed LAG, we conducted a ablation study. Results are presented in Figure ~\ref{fig:ablation}, with key observations as follows: The LLM-only baseline exhibited suboptimal performance in complex QA tasks. However, incorporating the retrieval module produced significant improvements, demonstrating the critical role of external knowledge retrieval. Specifically, adding the decomposition module further enhanced performance; we attribute to its ability to break down complex questions into simpler sub-questions, which facilitates more targeted and effective retrieval. Furthermore, integrating the reordering module led to additional gains by strengthening the logical coherence among sub-questions, optimizing the reasoning sequence. A more substantial performance boost was observed when introducing the core ``logical chain reasoning'' module, particularly in high-difficulty scenarios. This highlights the indispensable role of structured logical chains in guiding complex QA processes. Incorporating the atomic memory bank can further enhance the accuracy. Notably, incorporating the logical terminator module achieved the best overall performance. This improvement stems from its ability to mitigate error propagation in chain reasoning by terminating erroneous inference paths in a timely manner, thereby preventing cumulative errors. These findings confirm that each module adds unique value, necessitating their integration.

\section{Conclusion}
Existing RAG systems exhibit limitations in logical reasoning when addressing complex questions. Inspired by Cartesian principles, we propose LAG (Logic-Augmented Generation), a reasoning-first pipeline. The proposed adaptive decomposition module decomposes complex questions into atomic questions with logical dependencies. These atomic sub-questions are then solved sequentially following their logical dependencies via the proposed logical chain reasoning mechanism. Notably, we introduce a logical terminator mechanism that enables timely termination of the reasoning process when deviations occur, preventing error propagation in the logical chain and reducing wasted computation on low-value expansions. This framework perfectly aligns with the paradigm of solving complex questions based on Cartesian principles. Comprehensive experiments validate that proposed LAG outperforms conventional RAG systems in both multi-hop QA and challenging reasoning tasks, offering a principled alternative to existing RAG systems.

\section*{Limitations}
While LAG has advanced the paradigm for text-based retrieval-augmented generation, it currently lacks support for multimodal input sources. Extending this framework to incorporate multimodal inputs would enable a more comprehensive modeling of human information processing and reasoning mechanisms involving multimodal data. Given that real-world information is inherently multimodal, such an extension would further enhance the applicability and robustness of the LAG model.
\section*{Ethics Statement}
We confirm that we have strictly adhered to the ACL Ethics Policy throughout this study. Our research employs four publicly available datasets: HotpotQA, 2WikiMultiHopQA, MuSiQue, and GraphRAG-Bench. HotpotQA, 2WikiMultiHopQA, and MuSiQue are developed to assess the complex question-answering capabilities of various models, while GraphRAG-Bench is a domain-specific reasoning benchmark tailored for retrieval-augmented generation methods. All datasets utilized in this work have been widely adopted in RAG-related research and are free of private, sensitive, or personally identifiable information. We carefully selected these datasets to ensure compliance with ethical standards and to mitigate potential biases. Notably, our study does not involve the collection or modification of user-generated content, nor do we introduce synthetic data that may give rise to unintended misinformation.
\bibliography{custom}

\end{document}